\documentclass[preprint,12pt]{elsarticle}
\pdfoutput=1
\usepackage{amssymb}
\usepackage{amsmath}

\usepackage{multirow}
\usepackage{xcolor}
\usepackage{textcomp}
\usepackage{url}
\journal{Nuclear Physics B}

\begin{document}

\begin{frontmatter}



\title{Burst Image Super-Resolution via Multi-Cross Attention Encoding and Multi-Scan State-Space Decoding}

\tnotetext[1]{Funding: This work was supported by the National Natural Science Foundation of China under Grants 52127809 and 51625501.}

\tnotetext[2]{corresponding authors’ emails: liufulin@buaa.edu.cn, zhenzhongwei@buaa.edu.cn}

% \author{Tengda Huang}[style=chinese]

% \author{Yu Zhang}[style=chinese]

% \author{Tianren Li}[style=chinese]

% \author{Yufu Qu}[style=chinese]

% \author{Fulin Liu$^\ast$}[style=chinese]

% \author{Zhenzhong Wei$^\ast$}[style=chinese]


\author[t1]{Tengda Huang} %% Author name
% \thanks[t1]{main author}
\author[t1]{Yu Zhang}
% \thanks[t1]{}
\author[t1]{Tianren Li}
\author[t1]{Yufu Qu}
\author[t1]{Fulin Liu$^\ast$}
\author[t1]{Zhenzhong Wei$^\ast$}
\affiliation[t1]{organization={Instrumentation and Optoelectronic Engineering},%Department and Organization
            addressline={Beihang University}, 
            % city={Beijing},
            postcode={100080}, 
            state={Beijing},
            country={China}}

%% Abstract
\begin{abstract}
Multi-image super-resolution (MISR) can achieve higher image quality than single-image super-resolution (SISR) by aggregating sub-pixel information from multiple spatially shifted frames. Among MISR tasks, burst super-resolution (BurstSR) has gained significant attention due to its wide range of applications.
Recent methods have increasingly adopted Transformers over convolutional neural networks (CNNs) in super-resolution tasks, due to their superior ability to capture both local and global context. However, most existing approaches still rely on fixed and narrow attention windows that restrict the perception of features beyond the local field. This limitation hampers alignment and feature aggregation, both of which are crucial for high-quality super-resolution. To address these limitations, we propose a novel feature extractor that incorporates two newly designed attention mechanisms: overlapping cross-window attention and cross-frame attention, enabling more precise and efficient extraction of sub-pixel information across multiple frames. Furthermore, we introduce a Multi-scan State-Space Module with the cross-frame attention mechanism to enhance feature aggregation. Extensive experiments on both synthetic and real-world benchmarks demonstrate the superiority of our approach. Additional evaluations on ISO 12233 resolution test charts further confirm its enhanced super-resolution performance.
\end{abstract}


\end{frontmatter}



\section{Introduction}
\label{Intro}
%% Labels are used to cross-reference an item using \ref command.
Image super-resolution (SR) is a classic ill-posed inverse problem that aims to reconstruct a high-resolution (HR) image from one or more low-resolution (LR) inputs. SR methods can be broadly categorized based on the number of input LR images. 
Single-image super-resolution (SISR), which reconstructs a high-resolution (HR) image from a single low-resolution (LR) input, has been extensively studied and widely applied in various computer vision tasks, including security surveillance, medical image reconstruction, and video enhancement \cite{li2024systematic}. In contrast, multi-image super-resolution (MISR) leverages multiple LR images of the same scene to generate a higher-quality HR output, with applications such as burst photography \cite{dudhane_burstormer_2023}, remote sensing \cite{ye2024mra}, and video frame reconstruction \cite{kappeler_video_2016}. While most research efforts in recent years have primarily focused on SISR, the growing importance and practical demand for MISR in real-world applications call for increased attention and further investigation in this area.
The multiple LR images are captured by one camera with motion or multi cameras with offsets, thus there are sub-pixel shifts between multiple images. The core challenge lies in overcoming inevitable motions and effectively fusing features from multiple images, which often contain various degradations.
Therefore, existing MISR methods typically address the super-resolution task through the following steps: feature extraction, feature alignment, feature fusion, and high-resolution image reconstruction \cite{luo_ebsr_2021, luo_bsrt_2022, dudhane_burstormer_2023, cotrim_enhanced_2025}.

Compared with SISR, the MISR problems are able to aggregate sub-pixel information in multi-images, by alleviating the uncertainty problem in super-resolution \cite{wronski_handheld_2019}. MISR remains a less explored field than SISR, and the first deep learning-based approach \cite{kawulok_deep_2020} to generic MISR were reported in 2019. 

Some researchers handled MISR task using recurrent networks \cite{isobe_video_2020}, because they thought this task is closely related to video SR \cite{kappeler_video_2016}. However, such techniques of video SR are based on explicit or implicit assumptions related on input streams, such as the resolution can be increase by motion estimation when the fixed and rather high sampling frequency or presence of moving objects, and therefore are not applicable to all MISR programs. MISRGRU \cite{rifat_arefin_multi-image_2020} makes the information fusion using ConvGRU \cite{ballas_delving_2016}, which is based on a Recurrent Neural Network (RNN) architecture. The input of the model can be applied to a sequence of variable length. 
DeepSUM \cite{molini_deepsum_2020} is the first end-to-end MISR method, which extracts features from each LR image and estimates sub-pixel offsets to co-register feature maps, enabling direct application to unaligned inputs.

Most existing approaches extract features from each image independently and allow interaction only during alignment, leading to insufficient sub-pixel information fusion across images.
Some early methods use CNN-based architectures during the reconstruction phase, such as EBSR \cite{luo_ebsr_2021}. The long-range concatenation network was proposed for reconstruction in EBSR, it concatenates different level features for more global information. 
BIPNet \cite{dudhane2024burst} adds a global context attention mechanism in CNN structure to restore aligned features. 
Luo et al. introduced BSRT \cite{luo_bsrt_2022} by replacing the CNN-based backbone in EBSR with a standard Swin Transformer. While the 'small' version of the model achieves a notable reduction in parameter count, substantial performance improvements are primarily observed in the 'large' version, which incorporates a deeper stack of Swin Transformer blocks. However, this enhancement comes at the cost of increased computational complexity and hardware requirements. 
Burstormer \cite{dudhane_burstormer_2023} employs a no-reference feature enrichment module based on a cyclic burst sampling mechanism to enable inter-frame interaction while significantly reducing computational overhead. However, this design also introduces increased artifacts and blurring in the reconstructed images. 

\begin{figure}[!t]
\centering
\includegraphics[width=90mm]{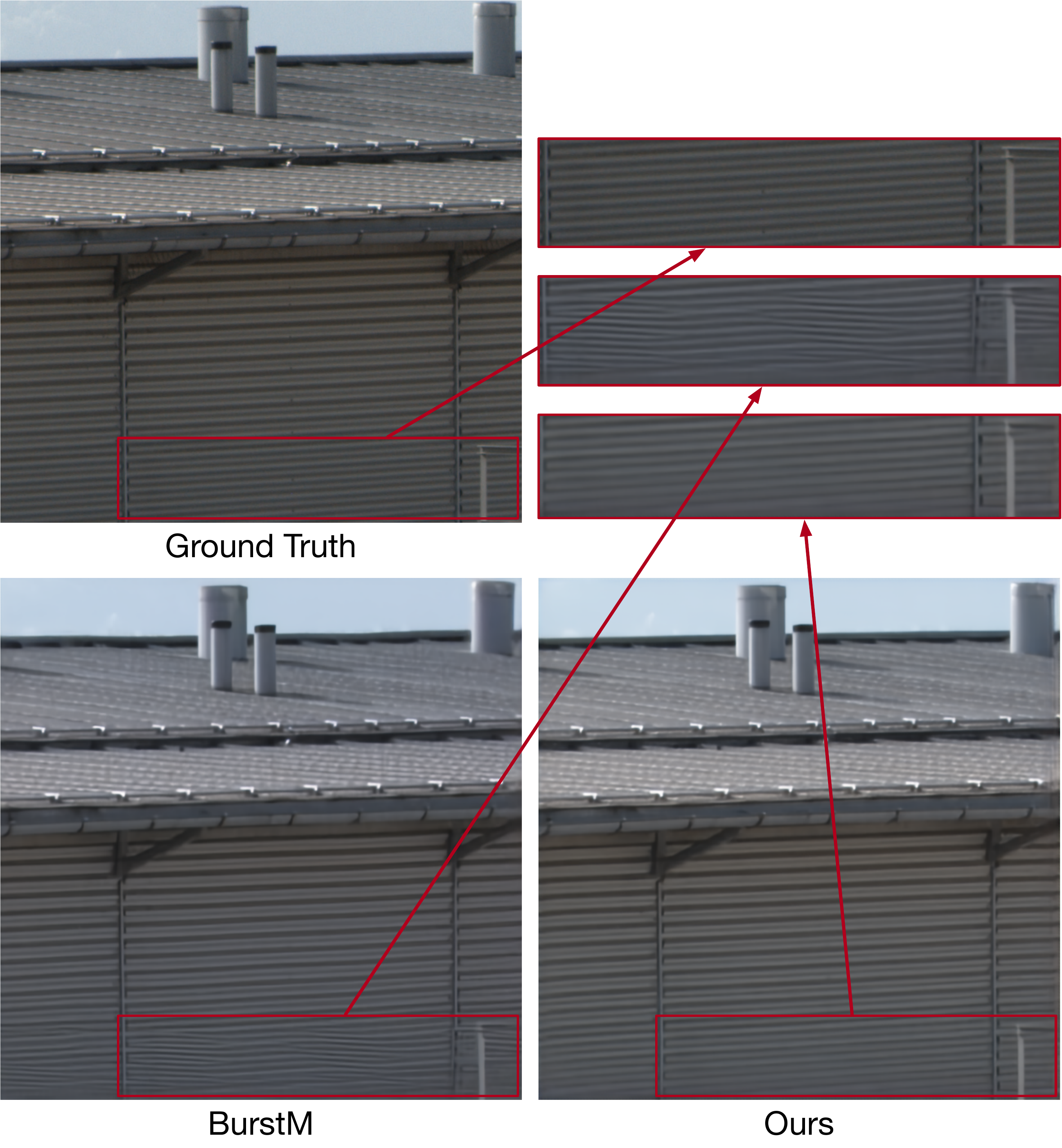}
\caption{Compared with the current SOTA method without DCN, our method shows better reconstruction results for real scenes.}
\label{fig_pre}
\end{figure}

While CNN-based architectures are efficient and suitable for deployment on resource-limited devices, their effective receptive field is inherently limited. Transformer based architectures provide a global receptive field by adjusting token granularity according to image resolution, yet their standard form introduces quadratic computational complexity that is often impractical.
Traditional Swin Transformer based architectures are constrained by fixed and narrow attention windows, which limits their ability to effectively sample features. As a result, sub-pixel information from burst images may be ignored if it falls outside the sampling window. Moreover, Swin Transformer based methods inherently face a trade-off between achieving a global receptive field and maintaining computational efficiency, often at the cost of reducing the effective receptive field. 
BurstM \cite{kang2024burstm} takes into account the limitations of deformable convolutional network (DCN), leading to misalignments between reference and source frames. They proposed the method using Fourier space with optical flow which conducts alignment using optical flow without DCN. However, this leads to reconstruction of distorted details that do not match the real scene when the optical flow estimate is not credible, as shown in Fig. \ref{fig_pre}. Recently, the structured SSM, as an effective high performance backbone, has been widely utilized in deep network construction \cite{fu_hungry_2023, gu_mamba_2024, gu_efficiently_2022, smith_simplified_2023}. This progress provides a potential solution for balancing the computational efficiency and global receptive field in image restoration. MambaIR \cite{guo_mambair_2024} takes advantage of Mamba in long-range modeling with linear complexity, and adapts Vision State-Space Module (VSSM) to image restoration. 
To better explore the sub-pixel information caused by inter-frame shifts, we focus on enhancing feature extraction and propose a Multi-Cross Attention mechanism for more effective information fusion.
We utilize DCN-based alignment to capture local spatial dependencies, and employ a Multi-Scan State-Space Module in the fusion stage. Multi-scan links features across pixel-level dithering ranges, while SSM ensures efficient long-range modeling, enhancing the overall feature integration.
In the feature alignment phase, we retain the DCN-based alignment that provides local constraints, and for the above limitations, we design Multi-Scan State-Space Module in the feature fusion phase to compensate for them. 
The multi-scan mechanism establishes connections among multiple scans within the dithering range of the target pixel. It further leverages SSM for long-range feature modeling with linear computational complexity, thereby enabling more effective feature fusion.

In short, the contributions can be summarized as follows:
\begin{itemize}
\item{
We propose a novel feature extractor, Multi-Cross Attention (MCA), which integrates overlapping Cross-Window Attention (CWA) and Cross-Frame Attention (CFA) to efficiently extract intra-frame features from multiple images.}
\item{
We propose a Multi-Scan State-Space Module (MS-SSM), which is jointly used with Cross-Frame Attention (CFA) to mitigate the misalignment introduced by DCN-based alignment, in the feature fusion stage.
}
\item{
We conduct comprehensive experiments on both synthetic and real-world datasets, demonstrating that our method consistently outperforms existing approaches. In addition, evaluations on the ISO 12233 resolution test chart further confirm the superior performance of our model in capturing fine-grained details.
}
\end{itemize}
% Section text. See Subsection \ref{subsec1}.

\section{Related Works}
\label{Related}
\subsection{Multi Image Super Resolution}
\label{Sub_MISR}
Multi image super resolution(MISR) can exploit the temporal redundancy among neighboring frames to achieve complementarity, yielding better performance compared with single image super-resolution (SISR) \cite{simonyan_very_2015, wang_real-esrgan_2021} that relies entirely on the nonlinear capability of deep network. Traditional MISR methods can be roughly divided into three categories: interpolation methods \cite{bose_superresolution_2006}, frequency-domain methods \cite{hui_ji_robust_2009} and regularization methods \cite{belekos_maximum_2010}. Unlike interpolation methods used in SISR, MISR typically involves three stages: registration, interpolation, and deblurring. Although these methods are simple and computationally efficient, their reconstruction results are often limited in quality. Frequency-domain methods utilize frequency transforms to extract high-frequency details of HR images, while regularization methods are particularly advantageous when dealing with a small number of LR images or significant blur. In recent years, deep learning-based methods have demonstrated significant advancements in MISR. 
Most deep learning-based MISR methods follow a standard pipeline consisting of feature extraction, alignment, fusion, and high-resolution image reconstruction.
DBSR \cite{bhat_deep_2021}, the first deep learning-based burst SR method, aligns frames using optical flow and employs attention-based strategies for fusion, achieving notable improvements in image reconstruction. 
MFIR \cite{bhat_deep_2021-1} builds upon DBSR by introducing improvements while continuing to use optical flow for frame alignment. It reconstructs high-resolution images through an advanced deep reparameterization formulation. 
% In \cite{kappeler_video_2016}, three consecutive motion-compensated video frames are processed by a deep CNN operating in both spatial and temporal domains. 这句视频超分的引用是否有点突兀
DeepSUM \cite{molini_deepsum_2020} uses shared 2D convolutions to extract high-dimensional features from each input image, and takes a slow fusion from these features by 3D convolutions. EBSR \cite{luo_ebsr_2021} combines feature-enhanced pyramid cascading with deformable convolution \cite{dai2017deformable} for multi-image alignment and employs cross non-local fusion to integrate features effectively. BSRT \cite{luo_bsrt_2022} introduces Swin Transformers and optical flow based on the EBSR framework, further increasing the network's complexity and achieving the NTIRE22 championship. BIPNet \cite{dudhane2024burst} proposes a pseudo-burst fusion strategy that enables frequent inter-frame interactions by fusing temporal features channel by channel. Meanwhile, Burstormer \cite{dudhane_burstormer_2023} enhances alignment and feature fusion through multi-scale local and non-local feature processing, further advancing the effectiveness of burst super-resolution methods. 
A more recent work introduced SBFBurst \cite{cotrim_enhanced_2025}, which accepts both RGB and RAW images as input.
In SBFBurst, two feature extraction modules are applied before and after the alignment module, and the method further leverages the mosaicked convolution technique introduced in MLB-FuseNet.
This approach enables the extraction of high-level features without disrupting the Bayer color pattern. However, it still inherits typical limitations of CNN-based methods, such as a restricted receptive field. 
BurstM \cite{kang2024burstm} argues that even multi-layer DCNs are insufficient for capturing global alignment due to their limited receptive field. To address this, it proposes a burst super-resolution method that performs alignment in the Fourier space using optical flow, evicting the use of DCNs.
However, the results of the reconstruction will be distorted in some real scenarios where the optical flow estimation results are not reliable, as shown in Fig. \ref{fig_pre}. 
Facing the limitations of DCN, in this paper, we use Multi-Scan State-Space Module method combined with CFA mechanism in the data fusion stage, and for the pixels of interest, we adopt 4 scans to search for the direction of the covering pixel that may produce displacement, and establish the SSM model. And CFA is utilized to establish the connection between the features obtained from SSM between pairs of frames. To solve the above problem.

\subsection{Vision Transformer}
\label{Sub_VTrans}
Recently, Transformer has gained popularity in the computer vision community due to its success in the field of natural language processing. Transformer based methods have demonstrated impressive performance in various high-level tasks, including image classification \cite{liu_swin_2021, dosovitskiy_image_2021, parmar_stand-alone_nodate}, object detection \cite{liu_swin_2021, vedaldi_end--end_2020, touvron_training_nodate, chu_twins_nodate}, segmentation \cite{wang_pyramid_2021, huang_glance_2022, cao_swin-unet_2021}, video inpainting \cite{liu_fuseformer_2021, zhang_exploiting_2024}. It has also been introduced for low-level vision tasks like image restoration \cite{luo_bsrt_2022, liang_swinir_2021, chen_pre-trained_2021}. 
Specifically, IPT \cite{chen_pre-trained_2021} developed a ViT-style network for various restoration problems based on the standard Transformer. However, IPT relies on patch-wise attention, which requires a large number of parameters. SwinIR \cite{liang_swinir_2021} addressed this issue by proposing an image restoration framework based on the Swin Transformer, reducing the parameter count while improving performance. BSRT \cite{luo_bsrt_2022} applied Swin Transformer to the multi-image super-resolution reconstruction task and achieved commendable results. 
Burstormer \cite{dudhane_burstormer_2023} explore inter-frame attention-based mechanisms to enhance feature interaction. They facilitate attention mechanism on the channel and spatial dimension to improve information exchange among frame. However, the fixed sampling window in spatial dimension attention makes it hard to mine valid information outside the window from burst images.
Despite these advancements, there is still space for us to design specifically for the MISR task.
Our approach addresses it by employing multi-cross attention, enabling better performance and more effective utilization of the Transformer architecture for burst super-resolution tasks.

\subsection{State Space Models}
\label{Sub_SSM}
State Space Models (SSMs) \cite{smith_simplified_2023, gu_combining_nodate} originated in control systems, where they are used for modeling continuous signal input systems. Recently, SSMs have been introduced into deep learning as a competitive backbone for state space transformation. Its favorable property of linear scalability with increasing sequence length in long-range dependency modeling has attracted significant interest from researchers. With advancements in SSMs, these models have also found applications in computer vision \cite{guo_mambair_2024, zhu_vision_2024, patro_simba_2024}. Notably, Visual Mamba incorporates a residual VSS module and introduces four scanning directions for visual images, achieving performance superior to that of ViT \cite{dosovitskiy_image_2021} while maintaining lower model complexity, thus receiving widespread recognition. In this work, we build upon the efforts of MambaIR  \cite{guo_mambair_2024} by applying a Mamba-based method to the multi-image super-resolution task. Our approach demonstrates improved performance over the use of the Transformer structure alone, further validating the potential of SSMs in enhancing image restoration tasks.

\section{Methodology} 

\begin{figure*}[!t]
\centering
\includegraphics[width=140mm]{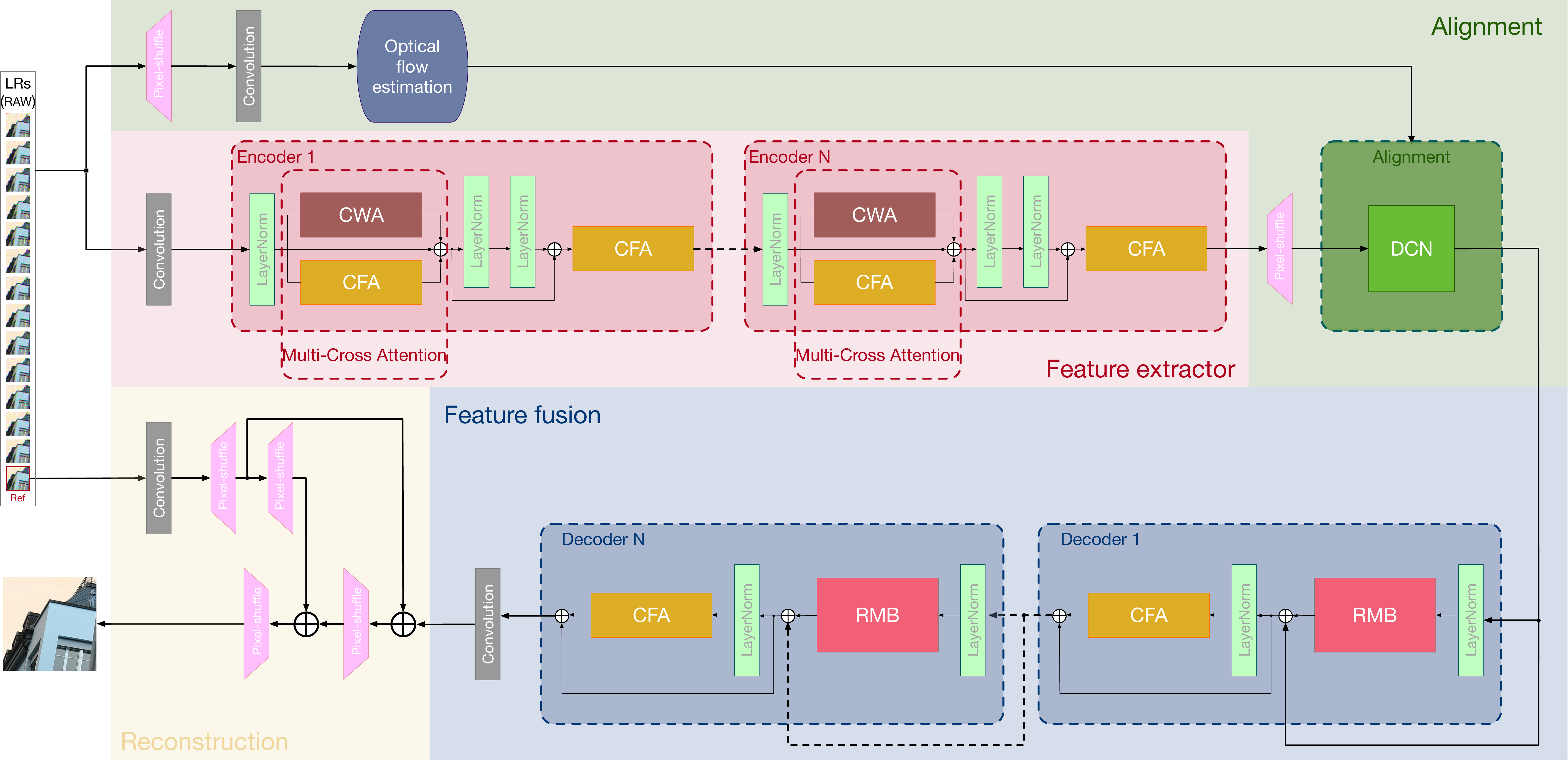}
\caption{
An overview of the proposed method. 
The network takes as input a burst of degraded RAW images and outputs a clean, high-quality sRGB image.
First, all RAW inputs are upsampled into an 'RGGB' format using PixelShuffle, and then expanded to 3 channels using a 3×3 convolution. 
These are subsequently fed into the optical flow estimation module to compute multi-scale flows between each frame and the reference. Meanwhile, features are directly extracted from the RAW inputs. Each RAW image is first projected into the feature space using a $3\times3$ convolution and then fed into our designed set of $N$ encoders. The resulting features are pixel-shuffled and aligned using the estimated optical flows. These aligned features are then processed by $N$ decoders for feature fusion, followed by residual upsampling to reconstruct the high-resolution image.
}
\label{fig_method_overview}
\end{figure*}

Our method aims to effectively leverage sub-pixel information in burst images for enhanced super-resolution reconstruction.
The overall framework of our proposed method is illustrated in Fig. \ref{fig_method_overview}. Analogous to Burstormer, we take $I_{HR} \in \mathbb{R}^{3 \times Hs \times Ws}$ as the ground truth high-resolution (HR) image, and the noisy raw burst images $\{x_i\}^N_{i=1}$ as the input. The model processes these inputs and produces a high-resolution RGB image, where $H$ and $W$ represent the height and width of the input image, $s$ is the scale factor, $N$ is the number of bursts, and $x_i \in \mathbb{R}^{4 \times \frac{H}{2} \times \frac{W}{2}}$ denotes the size of each burst image.

Initially, a convolutional layer, $F_{Conv}(.)$, is applied to map the raw burst images to the feature space $F_0 \in \mathbb{R}^{N \times C \times H \times W}$. Following this, we introduce a more efficient feature extraction module comprising $N$ Encoders (denoted as $E$), with particular emphasis on extracting sub-pixel information. The deeper features $F_d$ are subsequently extracted by the Encoders and forwarded to the Alignment module for further processing.
\begin{equation}
    F_d = E(F_0), F_d \in \mathbb{R}^{N \times C' \times\ H \times\ W}.
\tag{1}
\end{equation}

We retain the module based on deformable convolution network (DCN) for feature alignment. The offset estimation and feature alignment are performed using SpyNet \cite{ranjan_optical_2017} and multi-layer DCN. 
To overcome the limitations of DCN mentioned in BurstM, we introduce a Decoder (denoted as $D$) that combines Multi-Scan State-Space Model (MS-SSM) and Cross-Frame Attention (CFA). This decoder performs multi-directional scans to build spatial sub-pixel matching, addressing misalignment in uncertain directions during the alignment phase and enhance the feature aggregation.
The aligned features are then recovered into $F_a \in \mathbb{R}^{H \times W \times C'}$, and subsequently sent to $N$ decoders for feature fusion. We implement scaled-residual mechanisms between the $N$ decoders, leveraging long-range information dependencies with linear computational complexity. This facilitates the effective fusion of multi-frame information, enhancing the overall reconstruction performance.

\begin{equation}
    F_f = D(F_a) + s \cdot F_a,
\tag{2}
\end{equation}
where $s$ is the scale factor to control the feature residual. Finally, a single high-resolution RGB image is directly reconstructed in the reconstruction stage. 
We enhance the reference frame features in two stages using skip connections, following the application of PixelShuffle \cite{huang_multi_2009}.

\subsection{Feature Extraction}
\label{Method_FE}
The input raw images are first mapped to the feature domain $F_0 \in \mathbb{R}^{N \times C \times H \times W}$ through a $3 \times 3$ convolutional layer for shallow feature extraction.
This module utilizes enlarged cross-window attention and overlapping sampling to capture pixel information beyond the traditional sampling window caused by inter-frame displacement.

\begin{figure*}[!t]
\centering
\includegraphics[width=140mm]{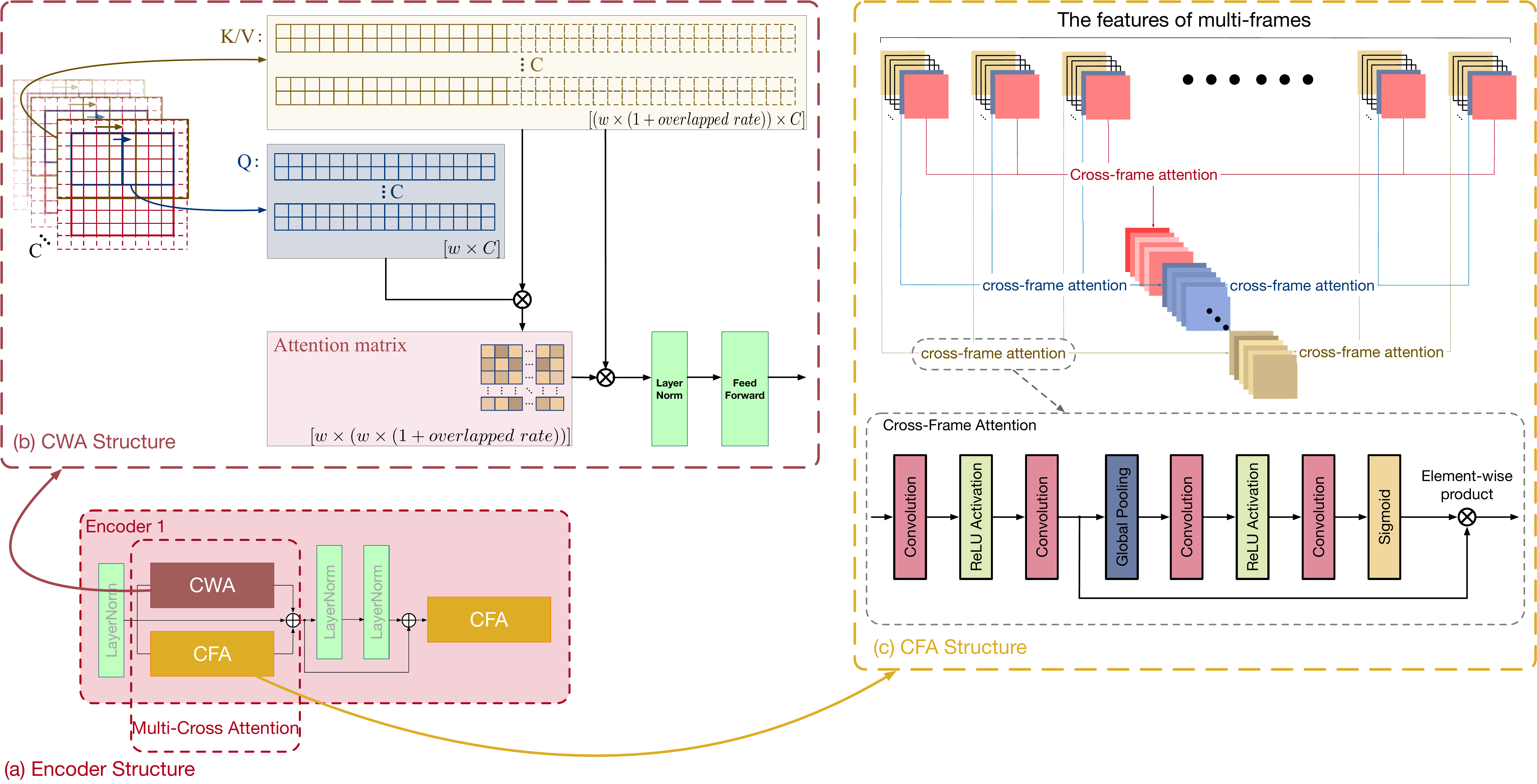}
\caption{Details of the proposed Encoder. Parallel Cross-Window Attention (CWA) and Cross-Frame Attention (CFA) are integrated to form a Multi-Cross Attention mechanism, with their detailed structures illustrated in Fig. (b) and (c), respectively.}
\label{Encoder_detail}
\end{figure*}

Through our observations, we found that when extracting features from a single low-resolution image, channel attention offers limited benefits to the extraction process while significantly increasing memory consumption. 
In contrast to channel attention, our proposed cross-frame attention mechanism facilitates more accurate extraction of sub-pixel information, which is critical to multi-frame super-resolution, with significantly reduced memory consumption.
The deep feature extraction stage consists of $N$ Encoders, each comprising a Cross-Window Attention (CWA) and a Cross-Frame Attention (CFA). The MCAB effectively focuses on both pixel-level displacements and sub-pixel information, enhancing the model's ability to capture fine-grained details. The resulting deep features, denoted as $F_d \in \mathbb{R}^{N \times C' \times H \times W}$, are extracted by the Encoder, where $C'$ represents the feature embedding dimension. The entire encoding process, which employs a residual structure, is formulated in Eq. \ref{Encoder}.
\begin{equation}
\begin{aligned}
    &F_{lw} = CWA(LN(F_0)) \\
    &F_{cf} = CFA(LN(F_0)) \\
    &F_{mc} = F_{lw} + \alpha F_{cf} + F_0 \\
    &F_d = MLP(LN(F_{mc})) + F_{mc}.
\end{aligned}
\label{Encoder}
\end{equation}
Within the encoder, the features first undergo layer normalization before being fed into the Cross-Window Attention (CWA) block and Cross-Frame Attention (CFA) block in parallel, as illustrated in Fig. \ref{Encoder_detail} (a).
In the CWA block, as shown in Fig. \ref{Encoder_detail} (b), cross-attention is computed within each window using pixel tokens. To enhance the receptive field, we apply cross-window attention by scanning the projected features with varying window sizes. Specifically, the larger, overlapping windows for the key (K) and value (V) features  allow attention to extend to pixels beyond the query ($Q$) window. 
For the input features $X_q, X_k, X_v \in \mathbb{R}^{H \times W \times C}$, we use larger window sizes for $X_k$ and $X_v$ compared to $X_q$. The $X_q$ features are divided into non-overlapping windows of size $P \times P$, resulting in ${HW}/P^2$ windows. In contrast, $X_k$ and $X_v$ are partitioned into overlapping windows of size $P' \times P'$, ensuring a broader context for attention. This process is formally defined in Eq. \ref{window size}. 

\begin{equation}
P' =  P \times (1 + r), \label{window size}
\end{equation}
where $r$ is overlapping rate to control the overlapping size. It is noteworthy that zero-padding with size $rP/2$ is applied to ensure consistent window sizes for overlapping windows. The attention matrix is calculated as Eq. \ref{attention matrix}. 
\begin{equation}
    Attention(Q,K,V) = SoftMax(QK^{T}/\sqrt{d}+B)V, \label{attention matrix}
\end{equation}
where $d$ is the dimension of $Q, K$ and $V$. $B$ is the learnable relative position encoding. 

Previous studies \cite{li_uniformer_2023, xiao2021early} have demonstrated that incorporating convolution into Transformer architectures can enhance visual representation learning and facilitate optimization. Inspired by this, we design our CFA block based on convolutional blocks, as illustrated in Fig. \ref{Encoder_detail}(c).
Since the multi-frame inputs produce a large number of feature channels, the resulting computational cost is substantial. To mitigate this, we reduce the number of channels in the first two convolutional layers by a factor of $\beta$. Specifically, the input features with $C$ channels are compressed to $C/\beta$ after the first convolutional layer, followed by activation. The features are then expanded back to $C$ channels through the second convolutional layer.
The global pooling operation is used to embed the frame-wise global spatial information into channels. As shown in Fig. \ref{Encoder_detail}(c), the sigmoid function works as a gating mechanism to fully capture frame-wise dependencies from the aggregated information by global pooling.

\subsection{Feature Alignment}
\label{Method_FA}
The lack of precise pixel-level displacement information poses a major challenge in burst super-resolution, despite the additional sub-pixel details introduced by camera motion. As a result, alignment remains a critical aspect of burst super-resolution.  
Recent methods, such as BurstM \cite{kang2024burstm}, perform alignment using optical flow on the Fourier space, but they struggle with reconstructing certain real-world scenes. To address this, we employ a combination of optical flow estimation and deformable convolution \cite{dai2017deformable}.

First, we feed $F_0$ into a pre-trained SpyNet \cite{ranjan_optical_2017} to compute the optical flows $f_i$ between each burst image and the reference image.
Due to random dithering, each frame may contain pixel displacements at varying scales. To capture these multi-scale offsets more effectively, we apply average pooling to generate three levels of optical flow $\{f_i^1, f_i^2, f_3^3 \}_{i=1}^N$.
For coarse alignment, the input features $F_d^i$ of each burst image are warped using the corresponding optical flows, producing the coarsely aligned features $F_i^w$, as defined in Eq. \ref{warp_1}.
\begin{equation}
    F_i^w = warp(F_d^i, f_i),
\label{warp_1}
\end{equation}
where $\text{warp}(\cdot)$ denotes the warping operator.
Next, $F_i^w$ is concatenated with the reference feature $F_d^{ref}$ and the corresponding flow $f_i$ to predict refined local offsets through convolution, as shown in Eq. \ref{warp_conv}. 
\begin{equation}
    O_i = L_{Conv}(F_i^w, F_d^{ref}, f_i),
\label{warp_conv}
\end{equation}
where $L_{\text{Conv}}$ represents the convolution layers.
Using these refined offsets, the features of multiple frames are aligned through a multi-layer deformable convolution network (DCN), as formulated in Eq. \ref{DCN}.
\begin{equation}
    F_i^A = DCN(F_d^i, O_i),
\label{DCN}
\end{equation}
where $F_i^A$ denotes the aligned feature of the $i-th$ frame.
To further enhance alignment accuracy, we adopt a three-level pyramid structure. From the third layer to the first, the predicted offsets and aligned features are progressively refined and propagated to the next layer through upsampling.

% \begin{figure*}[!t]
% \centering
% \includegraphics[width=140mm]{model_structures/Alignment_detail.pdf}
% \caption{Detailed architecture of the proposed Flow-Guided Deformable Convolutional Networks (FGDCN) Alignment.}
% \label{fig_Alignment}
% \end{figure*}

\subsection{Feature Fusion and Super-Resolution Reconstruction}
\label{Method_FF_R}
Inspired by the SSM's ability to capture long-range dependencies with linear computational complexity, we design the Decoder module. 

\begin{figure}[!t]
\centering
\includegraphics[width=140mm]{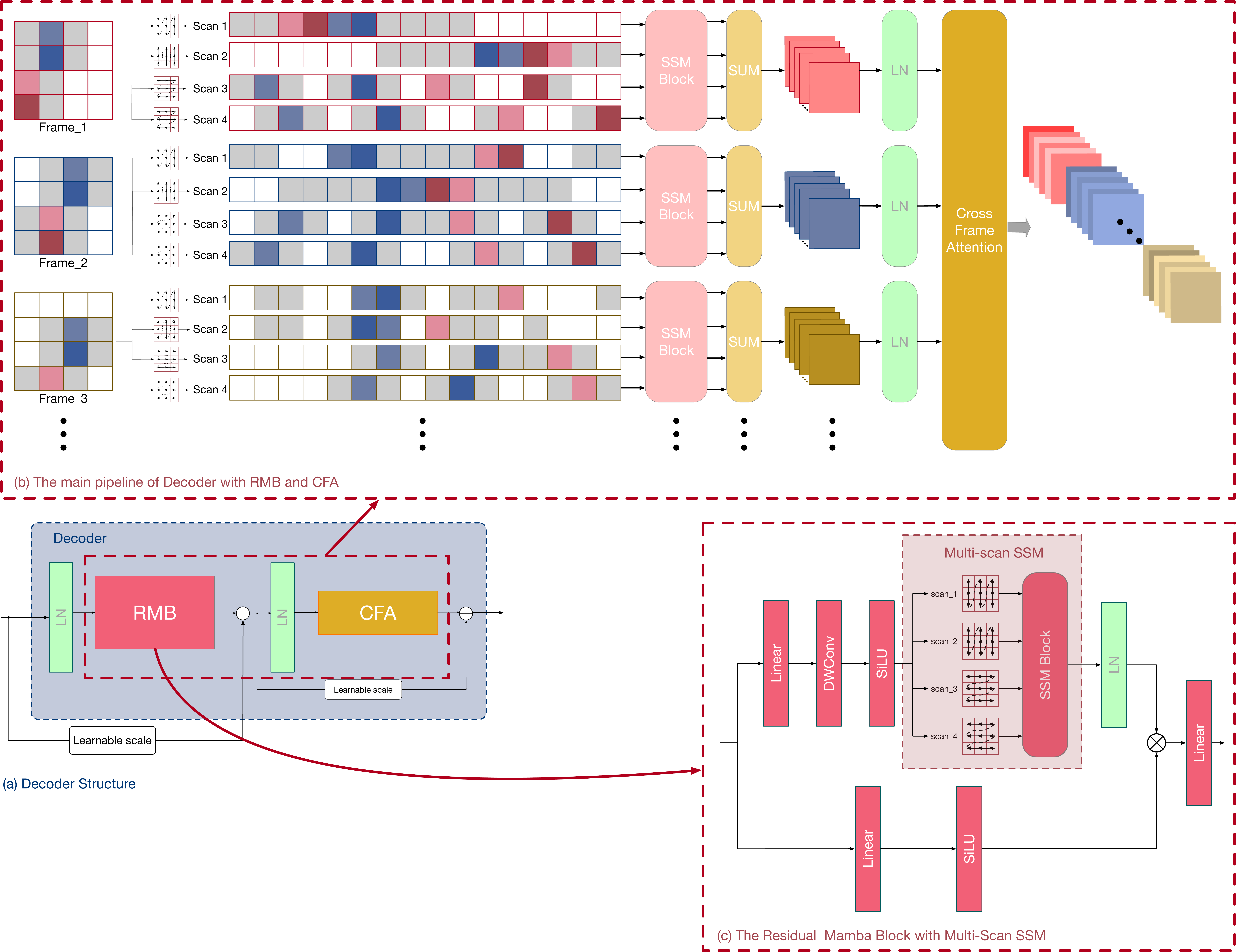}
\caption{Details of the proposed Decoder (a). The core pipeline is shown in (b). The designed Residual Mamba Block with Multi-Scan State-Space Module is shown in (c).}
\label{multi-scan_CFA}
\end{figure}

The decoder is composed of Residual Mamba Block (RMB) and Cross-Frame Attention (CFA), 
where a Multi-Scan State-Space Module (MS-SSM) designed for pixel displacement is the core of RMB.
As noted in BurstM \cite{kang2024burstm}, recent methods using multi-layer DCN with optical flow for alignment are insufficient. To address the spatial displacement differences between real-world and synthetic burst images, we enhance the Decoder module with MS-SSM and CFA, which further scan for potential misalignments, as illustrated in Fig. \ref{multi-scan_CFA}.
We first reshape the aligned deep features $F_N^A \in \mathbb{R}^{N \times H \times W \times C}$ into $NC \times H \times W$, then embed them into $C'$ dimensions using a convolutional layer. As formulated in Eq. \ref{Decoder}, given the input feature $F^A \in \mathbb{R}^{C' \times H \times W}$, spatial long-term dependencies are captured through layer normalization followed by an MS-SSM block. Additionally, we introduce a learnable scale factor $\gamma \in \mathbb{R}^{C'}$ to control the contribution of the skip connection.
\begin{equation}
    F_l = RMB(LN(F^A))+ \gamma \cdot F^A.
\label{Decoder}
\end{equation}

The RMB corrects misalignments and captures long-range dependencies using state space equations, as illustrated in Fig. \ref{multi-scan_CFA}(c). Following recent vision Mamba-based methods \cite{guo_mambair_2024, liu_vmamba_2024}, the input feature is processed through two parallel branches, as formulated in Eq. \ref{RMB}.
In the first branch, the feature channels are expanded to $\lambda C'$ using a linear layer, where $\lambda$ is a predefined channel expansion factor. The expanded features then pass through a depth-wise convolution (DWConv), the SiLU \cite{shazeer_glu_2020} activation function, the MS-SSM layer, and layer normalization.
In the second branch, the feature channels are also expanded to $\lambda C'$ but only processed by a linear layer followed by the SiLU activation. The outputs from both branches are then fused using the Hadamard product.
\begin{equation}
\begin{aligned}
    &F_1 = LN(MS-SSM(SiLU(DWConv(Linear(F))))) \\
    &F_2 = SiLU(Linear(F)) \\
    &F_{out} = Linear(F_1 \odot F_2),
\end{aligned}
\label{RMB}
\end{equation}
where $F_{\text{out}}$ is the final output of the RMB.
To restore local neighborhood similarity, we introduce a local convolution after the MS-SSM. Since the features are flattened into 1D token sequences in MS-SSM, nearby pixels in the 2D feature map may become distant in the 1D sequence, leading to potential local pixel forgetting. To mitigate this, we apply convolution layers to the normalized feature $F^{nor}$, using a bottleneck structure with a channel compression factor $s$. 
Additionally, we incorporate the CFA within the Decoder to enhance the representation of different frames, as shown in Eq. \ref{CFA_in_Decoder}. The input to the CFA is the sum of the RMB output and the skipped feature, as computed in Eq. \ref{Decoder}.
\begin{equation}
    F_{fused} = CFA(Conv(LN(F_l)))+ s' \cdot F_l,
\label{CFA_in_Decoder}
\end{equation}
where $s' \in \mathbb{R}^{C'}$ is a tunable scale factor controlling the residual connection.
Finally, the high-resolution image $I_{SR}$ is reconstructed using PixelShuffle \cite{huang_multi_2009}. Following common practices in super-resolution, we use the L1 loss between the restored image and the ground truth high-resolution image $I_{HR}$ as our objective function Eq. \ref{loss}.
\begin{equation}
    \mathcal{L} = ||I_{SR} - I_{HR}||
\label{loss}
\end{equation}

\section{Experiment And Analysis}
\label{Exp}
\subsection{Datasets}
\label{Data}
Following previous studies \cite{luo_ebsr_2021, luo_bsrt_2022, dudhane_burstormer_2023, bhat_deep_2021}, our method is evaluated on both synthetic and real-world datasets provided by the NTIRE2022 Burst Super-Resolution Challenge \cite{bhat2022ntire}. The synthetic dataset contains 46839 RAW burst sequences for training, which are used to generate low-quality images via random translations and rotations. High-resolution bursts are converted into low-resolution RAW burst sequences through downsampling, Bayer mosaicking, sampling, and the addition of random noise.

The real-world dataset comprises 5405 RAW burst patches, each measuring $160 \times 160$, captured using a Samsung Galaxy S8 smartphone. The corresponding high-resolution images were acquired using a DSLR camera. Additionally, 300 synthetically generated images (size $96 \times 96$) and 882 real-world patches ($160 \times 160$) are used for evaluation, with a $4 \times$ scaling factor.

The ISO 12233 resolution test chart, established by the International Organization for Standardization (ISO), is a standardized tool for assessing the resolution of digital cameras, lenses, and imaging systems. It quantifies visual resolution by analyzing specific patterns in the test chart. We developed an experimental platform using the ISO 12233 chart as a reference to evaluate the performance of our super-resolution method.

\subsection{Training details}
\label{Train}
The network was trained in a fully end-to-end manner while implicitly incorporating the demosaicing process. As is common practice, the network was initially trained on the synthetic dataset and subsequently fine-tuned on the real-world dataset. The input mosaicked RAW burst images were packed into a 4-channel RGGB format, while the outputs were in RGB format. These RGB outputs were further processed into visually interpretable images using the post-processing scripts provided in \cite{luo_bsrt_2022}.
For synthetic training, we used a pre-trained SpyNet with frozen weights for flow estimation. The network was optimized using the L1 loss and the Adam optimizer, with an initial learning rate of $1e^{-4}$, gradually reduced to $1e^{-6}$ via a cosine annealing schedule \cite{loshchilov_sgdr_2017}.
For training on real-world data, we employed an aligned L1 loss. As the ground truth images were not pre-aligned with the inputs, alignment was performed using a pre-trained PWC-Net \cite{sun_pwc-net_2018} to ensure accurate correspondence between the input frames and the ground truth images. The proposed network was implemented using the PyTorch framework and trained on 8 NVIDIA RTX 4090 GPUs.
In practice, we observed that using a larger patch size could further enhance performance. For synthetic training, we used a patch size of $384 \times 384$ for high-resolution images. However, due to memory constraints, the patch size for low-resolution images was reduced to \(64 \times 64\) during training on real-world data.

\begin{table*}[t]
\caption{The table shows a comparison between our methods and the other approaches on Synthetic and  real-world dataset for factor $\time 4$. 
\textcolor{red}{Red} and \textcolor{orange}{Orange} colors indicate the best and the second performance, respectively.
The values labeled '(64)' corresponds to the result of retraining on a real-world dataset using the same 64-patch size as ours, based on the open-sourced code from the paper.
The values for the Synthetic dataset and those marked with '(80)' are directly sourced from the respective papers; any omitted values indicate their absence in the original sources.
}
\centering
\label{tab:t1}
\scalebox{0.7}{
\begin{tabular}{|l|c|c|c|c|c|c|c|}
\hline
\multirow{2}*{Method} & \multirow{2}*{\#Parameters} & \multicolumn{3}{c|}{Synthetic dataset} & \multicolumn{3}{c|}{Real-world dataset} \\
\cline{3-8}
& & PSNR $\uparrow$ & SSIM $\uparrow$ & LPIPS $\downarrow$ & PSNR $\uparrow$ & SSIM $\uparrow$ & LPIPS $\downarrow$ \\
\cline{1-8}
BSRT \cite{luo_bsrt_2022}[2022] & 4.92M  & 42.72 & \textcolor{orange}{0.971} & \textcolor{orange}{0.031} & 48.48(80) & 0.985(80) & 0.021(80) \\

\cline{1-8}
\multirow{2}*{Burstormer \cite{dudhane_burstormer_2023}[2023]} & \multirow{2}*{3.58M} & \multirow{2}*{42.83} & \multirow{2}*{0.970} & \multirow{2}*{-} & 48.82(80) & 0.986(80) &  - \\
& & & & & 48.07(64) & 0.985(64) &  0.050(64) \\

\cline{1-8}
BIPNet \cite{dudhane2024burst}[2024] & 6.67M & 41.93 & 0.960 & 0.048 & 48.49(80) & 0.985(80) & 0.050(80) \\

\cline{1-8}
AFCNet \cite{mehta_adaptive_2022}[2024] & 47M & 42.21 & 0.960 & - & 48.63(80) & 0.986(80) & - \\

\cline{1-8}
\multirow{2}*{BurstM \cite{kang2024burstm}[2024]} & \multirow{2}*{14M} & \multirow{2}*{42.83} & \multirow{2}*{0.970} & \multirow{2}*{-} & 49.12(80) & 0.987(80) &  - \\
& & & & & \textcolor{red}{48.65(64)} & \textcolor{red}{0.986(64)} &  0.056(64) \\

\cline{1-8}
\multirow{2}*{SBFBurst \cite{cotrim_enhanced_2025}[2025]} & \multirow{2}*{7.64M} & \multirow{2}*{42.19} & \multirow{2}*{0.968} & \multirow{2}*{0.036} & 48.87(80) & 0.987(80) &  0.022(80) \\
& & & & & 48.06(64) & \textcolor{orange}{0.985(64)} &  \textcolor{orange}{0.022(64)} \\

\cline{1-8}
SeBIR \cite{liu2025sebir}[2025] & 8.25M & \textcolor{orange}{42.86} & 0.967 & 0.035 & 48.56(80) & 0.986(80) & 0.021(80) \\

\cline{1-8}
Ours & 12.60M & \textcolor{red}{43.20} & \textcolor{red}{0.972} & \textcolor{red}{0.028} & \textcolor{orange}{48.60(64)} & \textcolor{red}{0.986(64)} & \textcolor{red}{0.021(64)} \\
\cline{1-8}
\end{tabular}
}
\end{table*}

\subsection{Comparisons with Existing Methods}
\label{EXP_Compare}
We evaluate the proposed method on both synthetic and real-world datasets at a scale factor of $4 \times$, following the evaluation protocol of BSRT \cite{luo_bsrt_2022} and Burstormer \cite{dudhane_burstormer_2023}. Our method is compared with state-of-the-art BurstSR approaches, including BIPNet \cite{dudhane2024burst}, BSRT \cite{luo_bsrt_2022}, Burstormer \cite{dudhane_burstormer_2023}, BurstM \cite{kang2024burstm}, and SBFBurst \cite{cotrim_enhanced_2025}.
To ensure a comprehensive and robust comparison, we adopt PSNR, SSIM, and LPIPS as evaluation metrics. As shown in Table \ref{tab:t1}, the proposed method outperforms all other approaches on the synthetic dataset across all three metrics. In the real-world dataset, it achieves the highest scores for SSIM and LPIPS, while ranking second in PSNR.

\textbf{Results on the Synthetic BurstSR dataset.} 
As shown in Table \ref{tab:t1}, our method outperforms existing BurstSR methods, achieving state-of-the-art (SOTA) performance. Compared to the previous SOTA method, SeBIR, our approach yields a 0.34 dB improvement in PSNR and a 0.005 increase in SSIM. Additionally, our method also demonstrates further improvements in LPIPS, underscoring its effectiveness in enhancing perceptual image quality.

\begin{figure*}[!t]
\centering
\includegraphics[width=120mm]{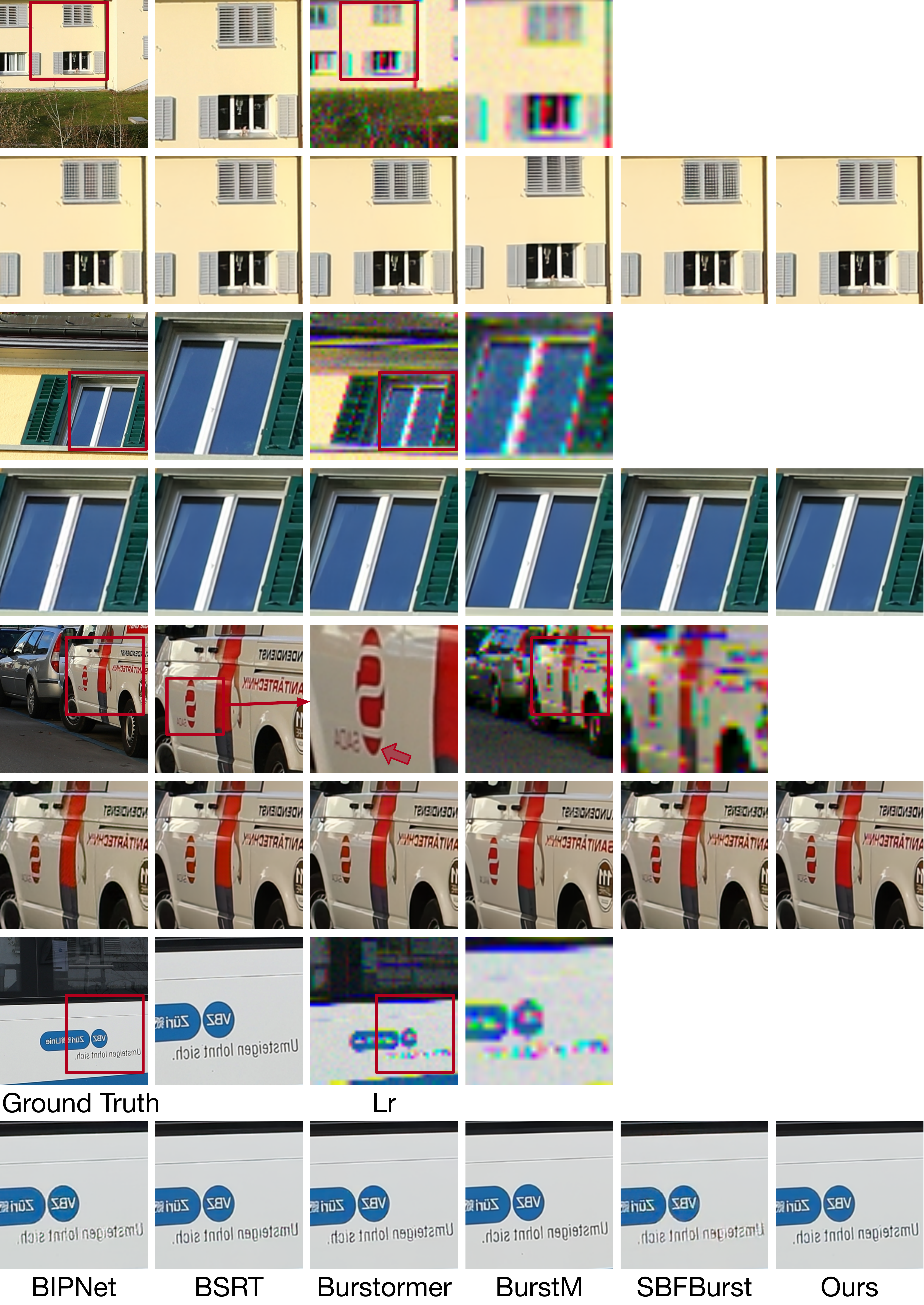}
\caption{
Visual comparison results on Synthetic Datasets \cite{bhat_deep_2021}.  The odd rows the Ground truth and the reference image in the input low-resolution images, as well as the corresponding zoom regions. The even rows depict the results of state-of-the-art methods.
}
\label{result_syn}
\end{figure*}

A visual comparison on the synthetic dataset is presented in Fig. \ref{result_syn}. We show representative examples highlighting text clarity, texture preservation, and structural accuracy. Compared to existing methods, our proposed approach demonstrates superior detail reconstruction and improved structural fidelity. This is evident in the first set of plots, where BSRT and BurstM produce noticeably blurred results, while BIPNet, Burstormer, and SBFBurst exhibit lattice artifacts that deviate from the ground truth. In the second set, our method reconstructs the green grid patterns more faithfully than all others. In contrast, BurstM and SBFBurst display pronounced distortions indicative of the “Jello effect.”
As discussed in the SBFBurst paper, many methods produce a darker red hue at the edges of the logo in the fourth set of images, particularly at the bottom, which differs from the ground truth. Upon close inspection of the ground truth, we find that the logo edges consist of dark pixels. The darker red reproduced by most methods stems from insufficient recovery of high-frequency detail. Notably, our method and BurstM achieve improved reconstruction quality in this region. Furthermore, for textual reconstruction, our approach shows higher consistency with the ground truth text patterns, as evident in the fourth and final examples.

\begin{figure*}[!t]
\centering
\includegraphics[width=100mm]{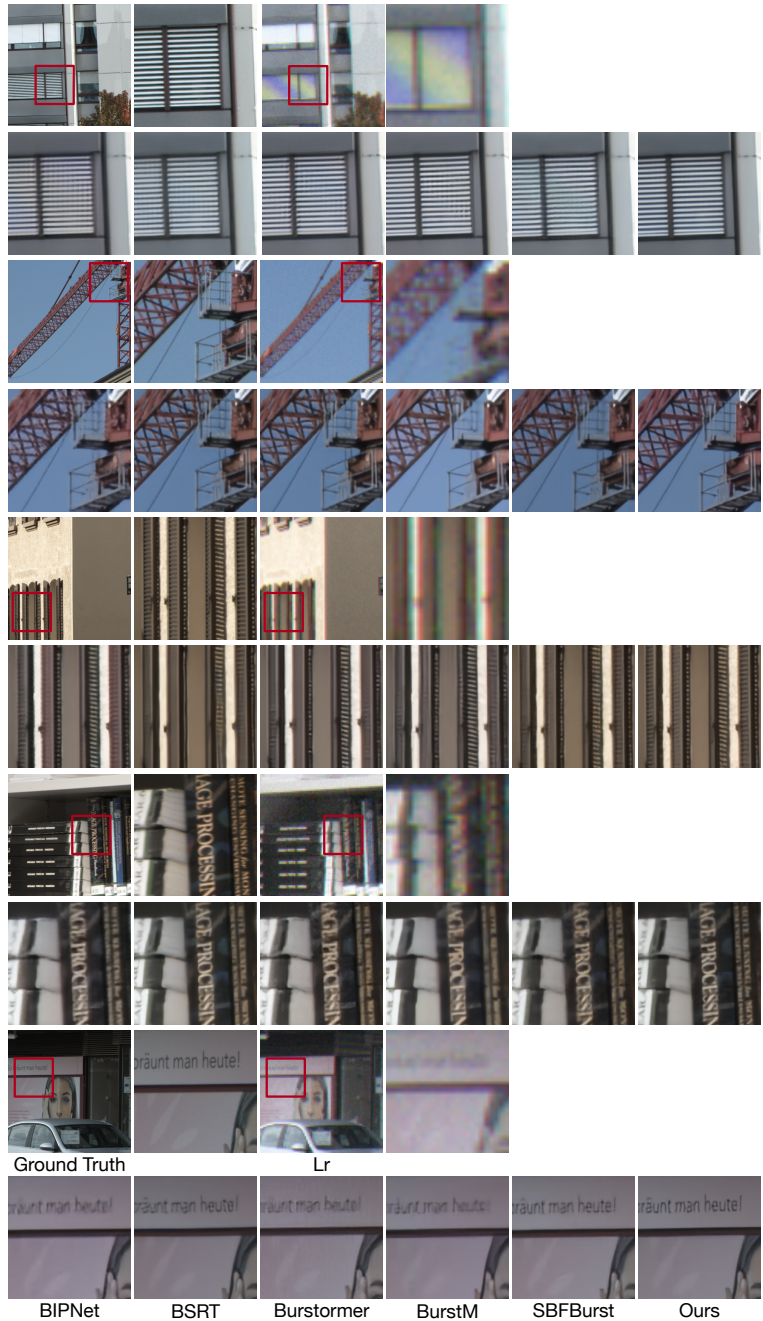}
\caption{
Visual comparison results on BurstSR Real World Datasets \cite{bhat_deep_2021}.The odd rows the Ground truth and the reference image in the input low-resolution images, as well as the corresponding zoom regions. The even rows depict the results of state-of-the-art methods.
}
\label{result_real}
\end{figure*}

\textbf{Results on the real-world BurstSR dataset.} 
In the real-world BurstSR dataset, there is a slight misalignment between the low-resolution (LR) bursts and high-resolution (HR), since they are captured using different cameras. Therefore, we trained our method using aligned L1 loss and evaluated it using aligned PSNR, SSIM, and LPIPS metrics, following the Burstormer protocol. The model pre-trained on the synthetic BurstSR dataset was fine-tuned for 100 epochs on the real-world dataset. As shown in Table \ref{tab:t1}, our method achieves a PSNR gain of 0.53 dB over the Burstormer. For LPIPS, our method shows a 0.029 improvement. Although our LPIPS score matches that of BSRT, it is worth noting that BSRT was trained using a patch size of 80, while our model was trained with a smaller patch size of 64.
Since larger patch sizes generally lead to better performance, we retrained several state-of-the-art methods for a fair comparison, including Burstormer \cite{dudhane_burstormer_2023}, BurstM \cite{kang2024burstm}, and SBFBurst \cite{cotrim_enhanced_2025}, on the real-world dataset. According to Table \ref{tab:t1}, AFCNet achieves PSNR and SSIM scores of 48.63 and 0.986, respectively, but it was trained with a patch size of 80. Remarkably, our method achieves competitive results despite being trained with a smaller patch size of 64. It outperforms both Burstormer and SBFBurst, which used the same training dimensions, and even surpasses SeBIR, which was trained with a larger patch size. Although BurstM achieves a higher PSNR, visual comparisons in Fig. \ref{fig_pre} reveal its qualitative limitations.

Visual results for real-world scenarios from the BurstSR dataset are presented in Fig. \ref{result_real}. In the first set of figures, our method demonstrates superior reconstruction of dense horizontal stripes on the shutters, while all other methods produce mesh-like artifacts; additionally, SBFBurst exhibits aliasing-like distortions. In the second set, depicting interlaced steel structures, our approach yields clearer reconstruction of the red metal framework and more accurate restoration of the gray metal guardrail, closely resembling the Ground Truth. Notably, none of the competing methods succeed in reconstructing the thin line within the red metal structure. In the third set, only our method successfully recovers the stripe patterns on the first and second window panes from the left, highlighting its superior ability to reconstruct fine details in small regions. In the fourth and fifth sets, the text restored by our method is noticeably sharper, while BIRNet, Burstormer, and BurstM generate more blurred results. Furthermore, in the fourth set, our method effectively reconstructs the horizontal lines formed by the book’s paper seams, which are poorly preserved by other methods.

\subsection{Resolution test chart experiment}
\begin{figure}[!t]
\centering
\includegraphics[width=3in]{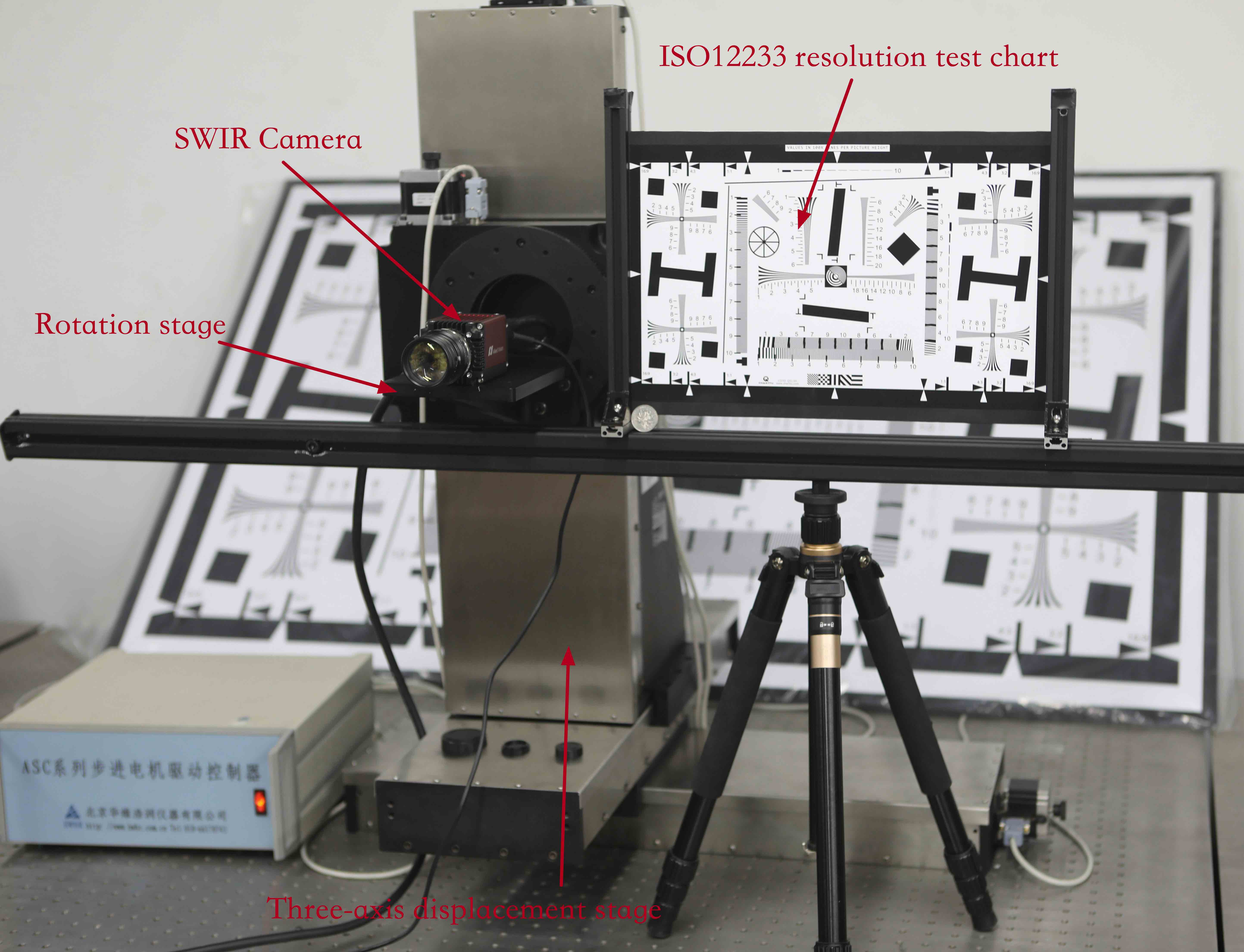}
\caption{Experimental setup for evaluating super-resolution methods using the ISO 12233 resolution test chart. A three-dimensional displacement stage with a short-wave infrared (SWIR) camera, and a resolution test chart mounted on a tripod.}
\label{expe_scene}
\end{figure}

We further evaluated our super-resolution method using the ISO 12233 resolution test chart as the reference target. To capture multiple images with sub-pixel shifts, we employed a short-wave infrared (SWIR) camera (InGaAs detector) mounted on a three-axis displacement stage. The camera was equipped with a 25 mm focal length lens from Grand Unified Optics. It features a pixel size of $15\mu m$ and a resolution of $640 \times 512$.
The imaging target was a $1\times$ ISO 12233 resolution test chart (100 LW/PH), measuring $177.8 \times 100mm^2$, mounted on a tripod-supported stand. The chart was illuminated by an LED light source emitting at a wavelength of $940nm$ (within the camera's spectral response range) to ensure that the brightest regions in the captured images approached the camera's maximum brightness capacity. This experimental setup facilitated a precise evaluation of our method’s capability to reconstruct fine image details. An illustration of the apparatus is provided in Fig. \ref{expe_scene}.

We collected 50 sets of data to evaluate our method. Each set comprised 14 images captured while the camera was randomly moved in directions parallel to the imaging plane. The object-to-camera distance was adjusted such that the test chart image fully spanned the horizontal field of view of the camera. The theoretical distance between the chart and the camera was 1.855 meters. When the chart was centered in the field of view (reference image), it was ideally aligned with the central region of the camera’s sensor. This configuration ensured consistent alignment and enabled an accurate assessment of the proposed method’s super-resolution performance.
To maintain controlled motion, we restricted the displacement between consecutive images to within 2 pixels. This value was computed based on the distance between the chart and the camera, as well as the intrinsic parameters of the imaging system.

\begin{figure*}[!t]
\centering
\includegraphics[width=90mm]{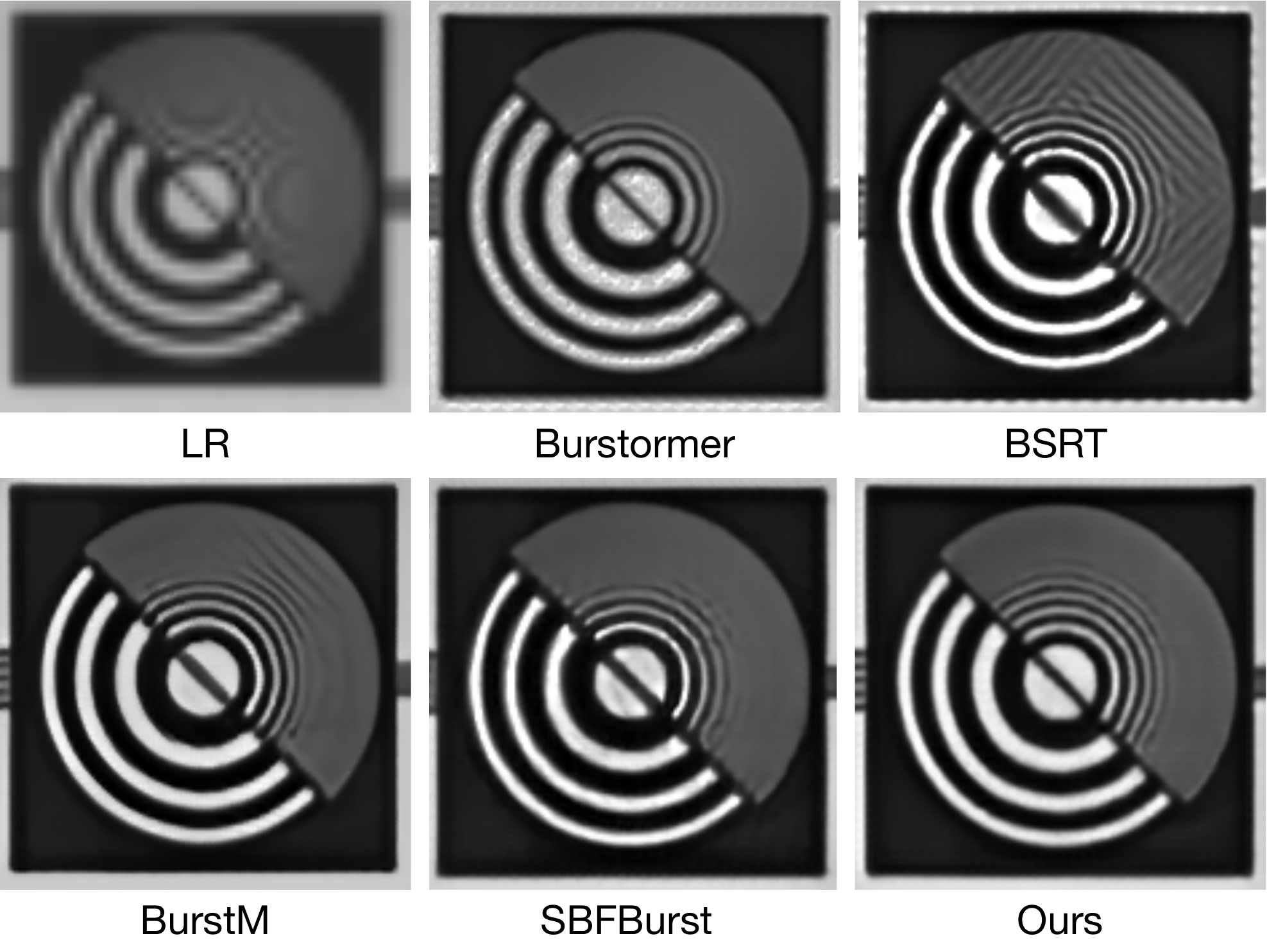}
\caption{Burst super-resolution results on the ISO 12233 resolution test chart (center). Shown are the original low-resolution (LR) image and the results of our method alongside state-of-the-art methods, zoomed into the central focus area. Our method reconstructs more circular stripe patterns with improved clarity.}
\label{chart_center}
\end{figure*}

\begin{figure*}[!t]
\centering
\includegraphics[width=90mm]{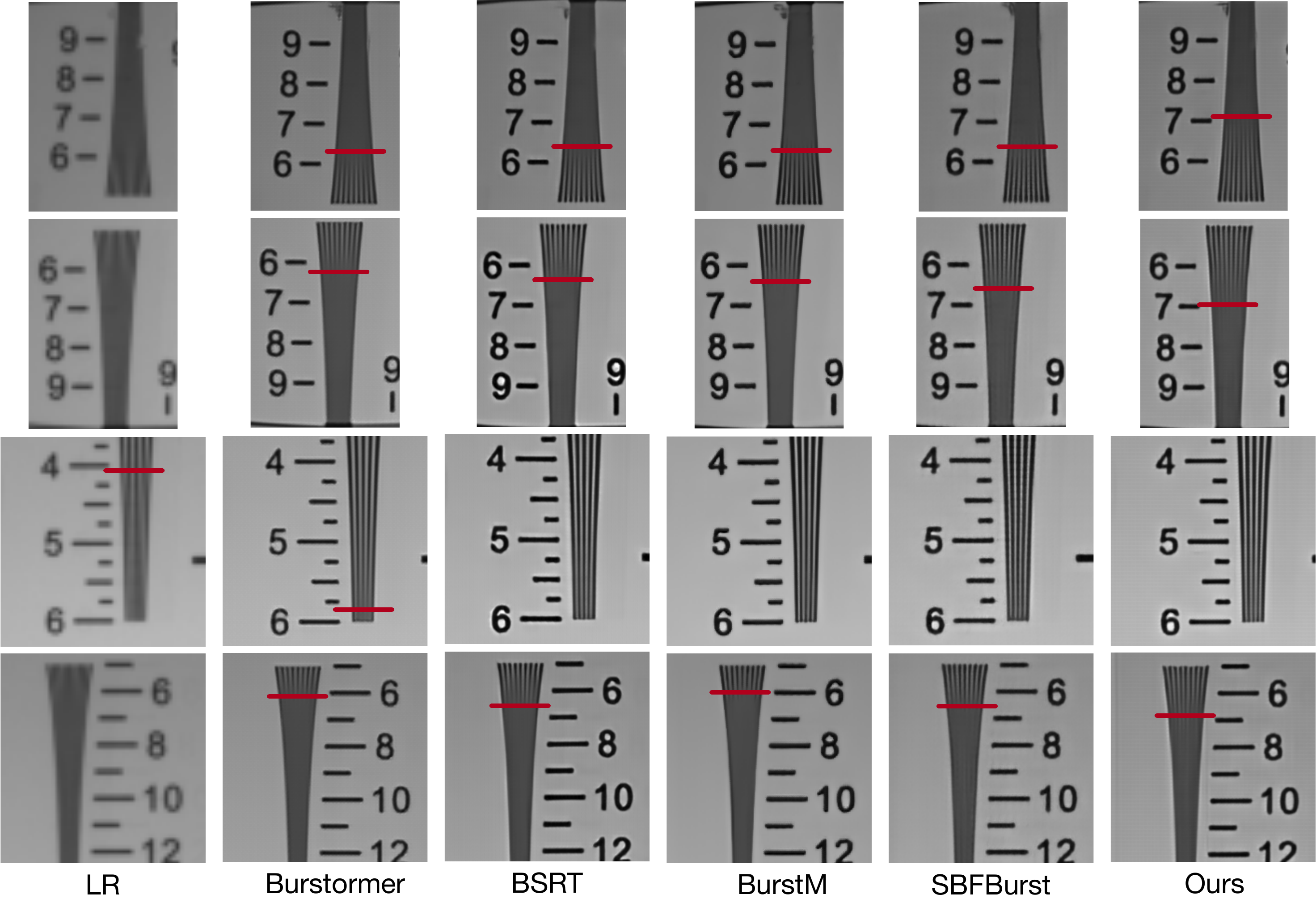}
\caption{
Burst super-resolution results on the ISO 12233 resolution test chart (horizontal). The original low-resolution (LR) image and the reconstruction results of our method and other state-of-the-art methods are shown, zoomed into the horizontal resolution area of the TV lines. Our method demonstrates superior horizontal resolution in the reconstructed output.
}
\label{chart_horizontal}
\end{figure*}

\begin{figure*}[!t]
\centering
\includegraphics[width=90mm]{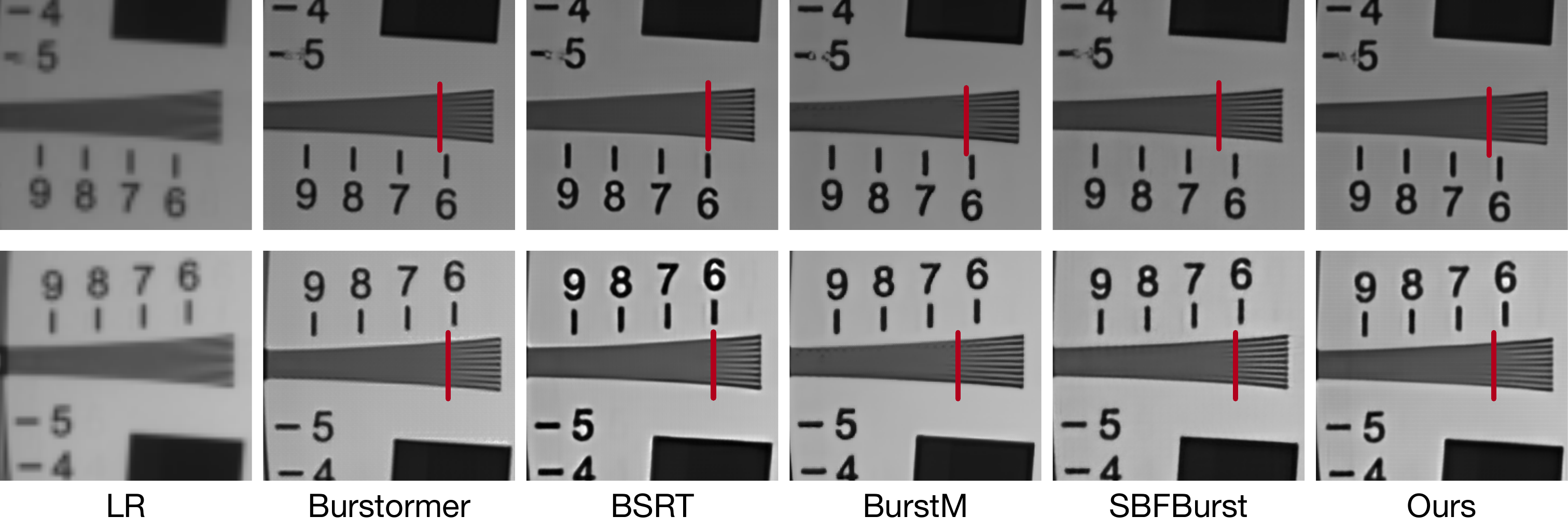}
\caption{
Burst super-resolution results on the ISO 12233 resolution test chart (vertical). The original low-resolution (LR) image and the reconstruction results of our method and other state-of-the-art approaches are shown, zoomed into the vertical resolution region of the TV lines. Our method achieves higher vertical resolution in the reconstructed results.
}
\label{chart_vertical}
\end{figure*}

\begin{figure*}[!t]
\centering
\includegraphics[width=90mm]{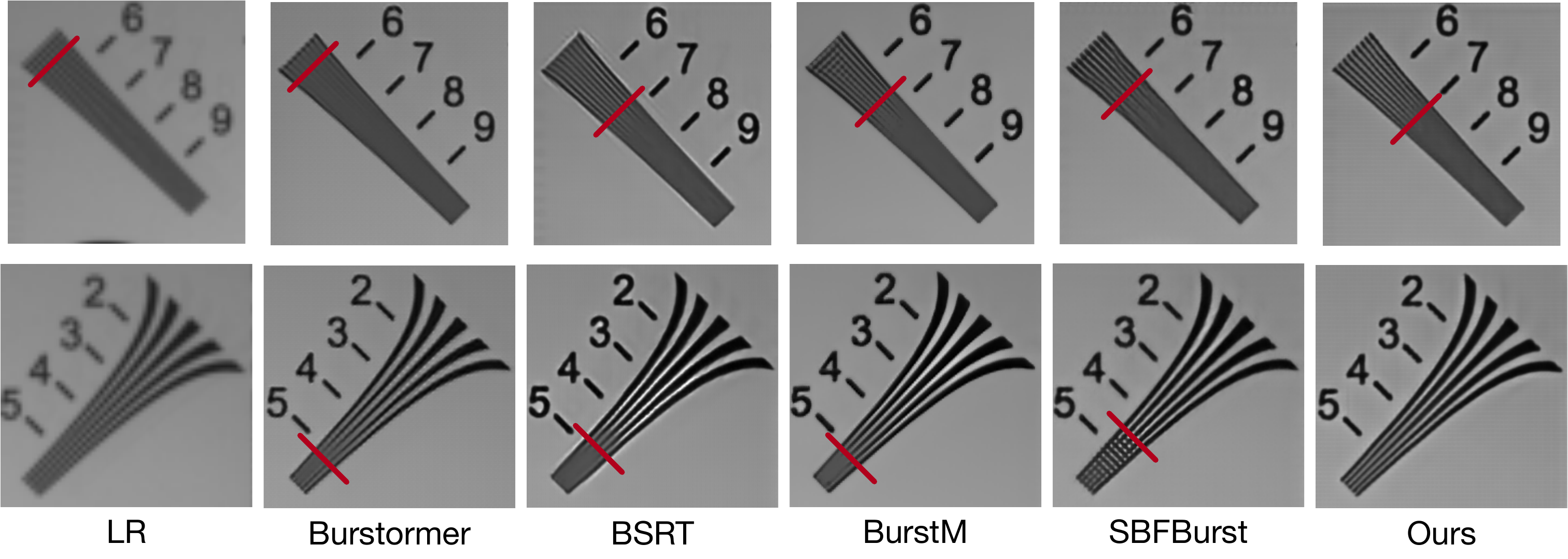}
\caption{
Burst super-resolution results on the ISO 12233 resolution test chart ($45^{\circ}$). The original low-resolution (LR) image and the reconstruction results of our method and other state-of-the-art approaches are shown, zoomed into the $45^{\circ}$ resolution region of the TV lines. Our method achieves higher reconstruction accuracy in this diagonal direction.
}
\label{chart_45}
\end{figure*}

The low-resolution images were upsampled to high-resolution counterparts (with a $4\times$ scaling factor) using several state-of-the-art BurstSR methods for comparative evaluation. The image resolution was directly assessed by examining specific regions of the ISO 12233 resolution test chart and analyzing contrast variations, which serve as indicators of reconstruction quality.
As shown in Fig. \ref{chart_center}, our method reconstructs a clearer and more distinguishable central focus area. 
Similarly, in Figs. \ref{chart_horizontal}, \ref{chart_vertical}, and \ref{chart_45}, the results produced by our method outperform those of Burstormer and BSRT. 

Given that 360 pixels correspond to a sensor height of 5.4 mm, the resolution measurements in LW/PH can be converted to LP/mm using a conversion factor of 9.26 for direct comparison.
After conversion, as shown in Fig. \ref{chart_horizontal}, our method achieves a horizontal resolution of 66 LP/mm, while Burstormer and BSRT reach 57 LP/mm and 59 LP/mm, respectively. BurstM and SBFBurst do not show significant improvements in this region.
In the vertical direction, shown in Fig. \ref{chart_vertical}, our method achieves a resolution of 59 LP/mm, surpassing both Burstormer and BSRT, which reach 56 LP/mm. Although BurstM and SBFBurst yield slightly better results in one specific instance, BurstM introduces considerable artifacts, undermining its visual quality.
In the $45^\circ$ directional resolution (Fig. \ref{chart_45}), our method attains 65 LP/mm, compared to 63 LP/mm by BSRT and a lower value from Burstormer. In this case, our method's advantage is more evident, as both BurstM and SBFBurst introduce noticeable artifacts.
These results collectively demonstrate the superior performance of our method across different resolution directions.

\subsection{Ablation Study}
\label{Sub_AS}
\begin{figure}[!t]
\centering
\includegraphics[width=90mm]{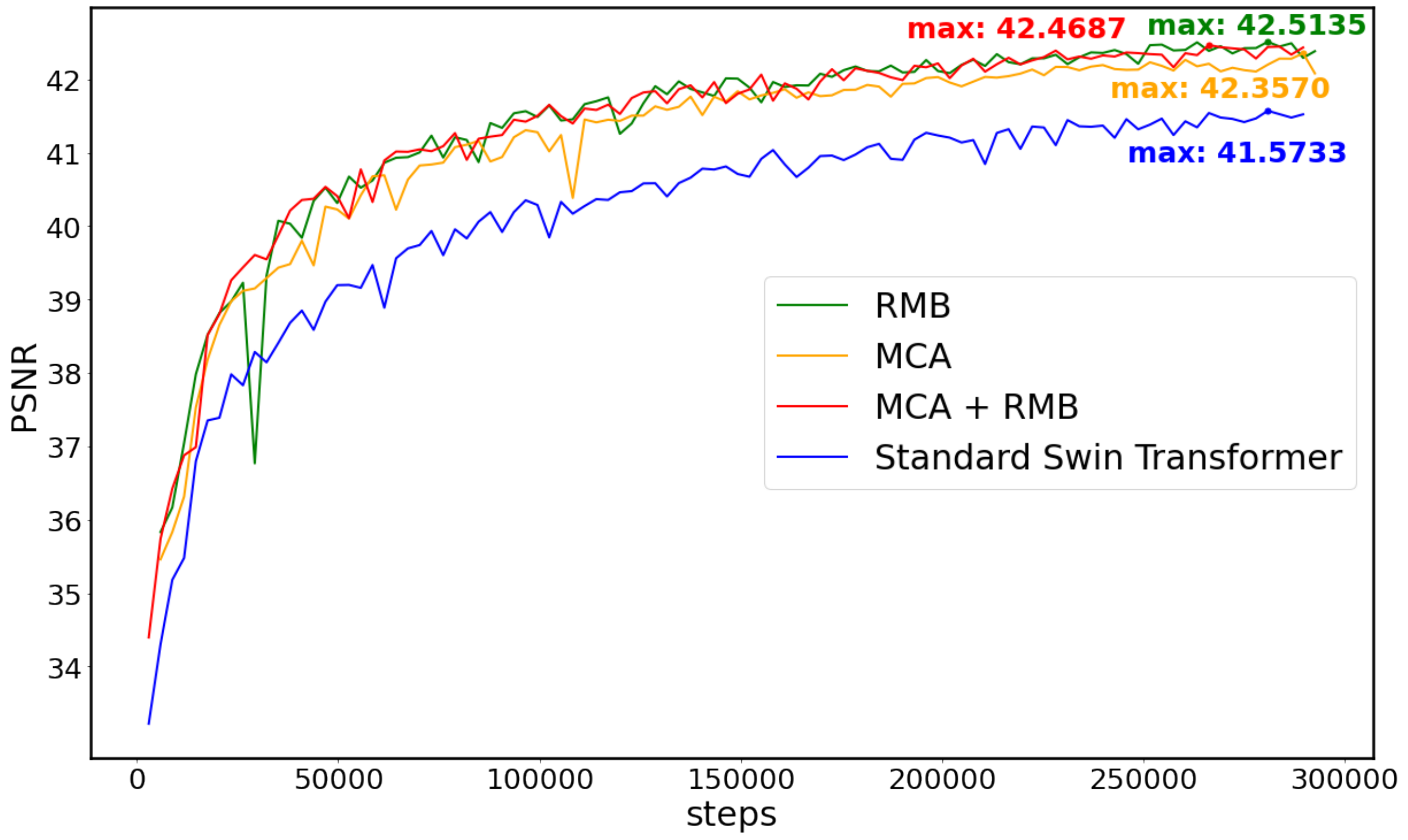}
\caption{
Comparison of training processes among the baseline method (Standard Swin Transformer), the model with only RMB, the model with only MCA and the full proposed model (MCA + RMB).
}
\label{fig_ablation}
\end{figure}

In this section, we demonstrate the effectiveness of individual modules within the proposed architecture. The original BSRT, which utilizes the standard Swin Transformer for both feature extraction and reconstruction, is adopted as the baseline model (represented by the blue line in Fig. \ref{fig_ablation}) for comparison. Multiple ablation models were trained on the SyntheticBurst dataset for 100 epochs, with their training processes illustrated in Fig. \ref{fig_ablation}.
The orange line in Fig. \ref{fig_ablation} represents the ablation results obtained using the proposed MCA modules in the decoder for feature extraction. Compared to the baseline with the standard Swin Transformer, the MCA modules yield a significant PSNR improvement of 0.78 dB.
The red line illustrates the performance achieved by integrating both the MCA feature extraction module and the RMB fusion module in the decoder. This configuration further improves the PSNR by an additional 0.11 dB and contributes to a more stable training process.
These results validate the complementary and synergistic advantages of incorporating the proposed modules into the network architecture.

The proposed RMB module facilitates more effective utilization of long-range global dependencies while maintaining linear computational complexity. 
The green line in Fig. \ref{fig_ablation} represents the model that incorporates only the RMB module. During the initial 100 training epochs, the performance of this model is comparable to that of our full method, indicating no significant advantage for the latter in the early training phase.

\begin{figure}[!t]
\centering
\includegraphics[width=90mm]{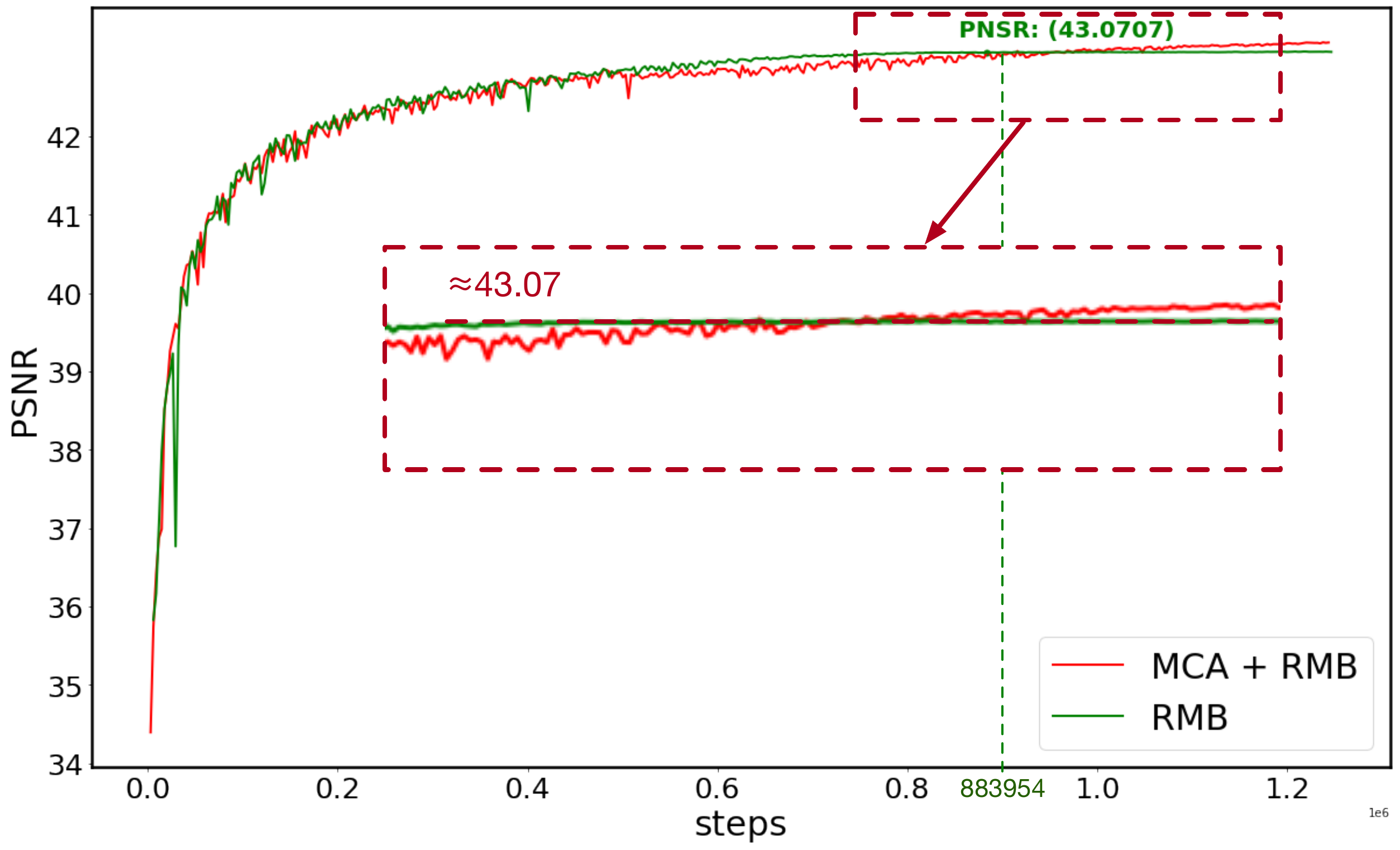}
\caption{Comparison between the training processes of the model with only RMB and our full model (MCA + RMB).}
\label{fig_ablation_mamba}
\end{figure}

As shown in Fig. \ref{fig_ablation_mamba}, the RMB reconstruction module alone demonstrates a faster convergence rate during the early stages of training, achieving performance that approaches or even temporarily surpasses that of the complete model (MCA + RMB). However, due to the limited representational capacity of its feature extraction module, the model ultimately converges around a PSNR of 43.07. 
In contrast, our complete model, which integrates both the MCA feature extraction module and the RMB reconstruction module, continues to improve in the later training stages, ultimately achieving superior overall performance.

\section{Conclusion}
\label{Conclusion}
We present a multi-image super-resolution method based on Transformer and Mamba. 
We designed new encoder and decoder for feature extraction and feature aggregation in the Burst super-resolution task. In the encoder, we proposed a new multi-cross attention architecture combines cross-window attention and cross frame attention, which could pay more attention to the inter-frame displacement and sub-pixel information. 
Moreover, we identify the limitations of the current state-of-the-art method, BurstM, which highlights the drawbacks of DCN-based alignment. In contrast, our approach retains DCN-based alignment to preserve its local constraint benefits, while addressing its limitations through a carefully designed decoder. Specifically, we introduce a Multi-Scan State-Space Module (MS-SSM) within the decoder, which is combined with a Cross-Frame Attention block to enhance feature aggregation. This design enables the model to better handle misalignment issues during feature fusion, which are often caused by the limited receptive field of DCNs.

Our results on synthetic datasets demonstrate that our method achieves a state-of-the-art performance. In the resolution chart experiments, we compared our method with current state-of-the-art approaches which based on deep learning. The results, obtained using the same training data, demonstrate that our method is more effective. 
On the real-world dataset, our method achieves performance comparable to current state-of-the-art approaches in terms of PSNR and SSIM, while producing noticeably fewer visual artifacts. In the resolution chart experiments, we further compare our method with existing deep learning-based techniques using the same training data. The results demonstrate that our method provides more effective detail restoration and superior perceptual quality.

\bibliographystyle{elsarticle-num}

\bibliography{reference}

\end{document}